\pdfoutput=1

\documentclass[11pt]{article}

\usepackage{ACL2023}

\usepackage{times}
\usepackage{latexsym}
\usepackage{amsmath}
\usepackage{hyperref}
\usepackage[T1]{fontenc}

\usepackage[utf8]{inputenc}

\usepackage{microtype}
\usepackage{graphicx}
\usepackage{kky}
\usepackage{booktabs}
\usepackage{multirow}
\usepackage{float}
 
\usepackage[textsize=scriptsize]{todonotes}

\usepackage{cleveref}
\newcommand{\secref}[1]{\S\ref{#1}}
\usepackage{enumitem}
\usepackage[hang,flushmargin]{footmisc}

\usepackage{mfirstuc}
 \newcommand{\ssc}[1]{\textsc{\capitalisewords{\MakeLowercase{#1}}}}

\usepackage{xspace}
\newcommand{\embedless}{\ssc{ONE-HOT}\xspace}
\newcommand{\dense}{\ssc{DENSE}\xspace}
\newcommand{\conv}{\text{conv}}
\newcommand{\repeatt}{\text{repeat}}
\newcommand{\byteconv}{Byte-$n$CF\xspace}
\newcommand{\byteword}{Byte-WSF\xspace}

\usepackage{subcaption}
\title{Local Byte Fusion for Neural Machine Translation}

\author{Makesh Narsimhan Sreedhar$^1$, Xiangpeng Wan$^2$, Yu Cheng$^3$, Junjie Hu$^1$ \\
$^1$University of Wisconsin-Madison, $^2$NetMind.AI and ProtagoLabs, $^3$Microsoft Research \\
\texttt{\{msreedhar,junjie.hu\}@wisc.edu}
}

\begin{document}
\maketitle
\begin{abstract}
Subword tokenization schemes are the dominant technique used in current NLP models. However, such schemes can be rigid and tokenizers built on one corpus may not adapt well to other parallel corpora. It has also been observed that in multilingual corpora, subword tokenization schemes oversegment low-resource languages, leading to a drop in translation performance. An alternative to subword tokenizers is byte-based tokenization, i.e., tokenization into byte sequences using the UTF-8 encoding scheme. Byte tokens often represent inputs at a \textit{sub-character} granularity, i.e., one character can be represented by a span of byte tokens. This results in much longer byte sequences that are hard to interpret without aggregating local information from multiple byte tokens. 
In this paper, we propose a \textbf{Lo}cal \textbf{B}yt\textbf{e} \textbf{F}usion (LOBEF) method for byte-based machine translation---utilizing byte $n$-gram and word boundaries---to aggregate local semantic information. Extensive experiments on multilingual translation, zero-shot cross-lingual transfer, and domain adaptation reveal a consistent improvement over vanilla byte-based models. Further analysis also indicates that our byte-based models are parameter-efficient and perform competitive to subword models.   
\end{abstract} 

\section{Introduction}
\label{sec:introduction}

\begin{figure*}
    \centering
    \includegraphics[width=0.9\textwidth, height=4.5cm]{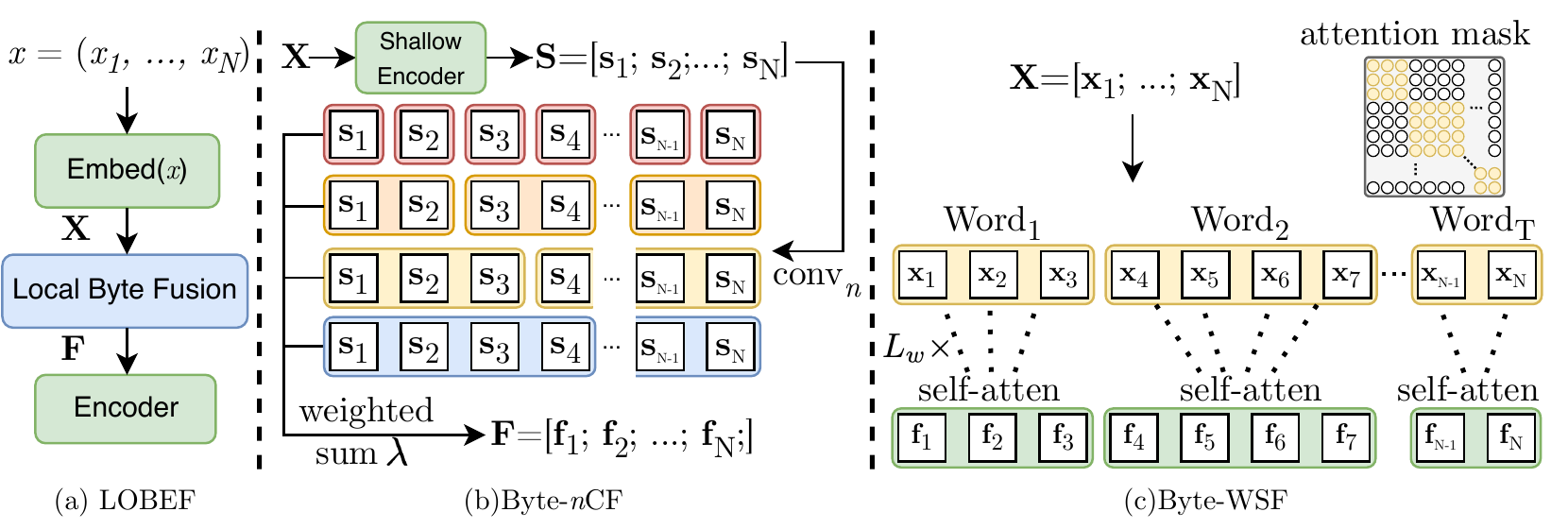}
    \vspace{-4mm}
    \caption{(a) LOBEF; (b) \textbf{\byteconv} uses four convolutional layers (width=$n$, stride=$n$) to aggregate char-level information; (c) \textbf{\byteword} uses word boundaries with block-wise self-attention to aggregate word-level information.}
    \vspace{-3mm}
    \label{fig:main-figure}
\end{figure*}
Multilingual neural machine translation (NMT) has proven effective to transfer knowledge learned from a high-resource language to a low-resource language.
However, existing multilingual NMT models still rely on a pre-built subword tokenizer (e.g., BPE~\citep{sennrich-etal-2016-neural}, SentencePiece~\citep{kudo-richardson-2018-sentencepiece}) to tokenize a sentence into a sequence of subword units. This has two drawbacks. First, once the tokenizer is fixed, we lose the flexibility of changing the word tokenization if we aim to fine-tune the NMT model on another parallel corpus of interest for adaptation. Second, when a subword tokenizer is built on unbalanced multilingual data, word tokens from a low-resource language are usually under-represented, resulting in over-segmentation of a word into many single characters. A recent study~\citep{rust-etal-2021-good} measures over-segmentation by the \textit{fertility} score of a subword scheme which indicates how many subwords a whole word is broken down into. It shows a negative correlation between the fertility score and the performance of multilingual models on the over-segmented languages. 


Although character-based models have often been proposed as a solution to these problems \citep{gupta-character-transformer, libovicky-fraser-2020-towards, li-etal-2021-char}, they come with their own tradeoffs related to the significant overhead of processing long character sequences during training and inference \citep{libovicky-fraser-2020-towards}. Besides, these models still adopt a fixed vocabulary of characters, leading to the same issue as a fixed subword tokenizer for adaptation. Another point of difference is that in character-based methods, the vocabulary still consists of one unique embedding for each character under consideration. In byte-based approaches, the tokens are at a sub-character granularity and the model has to figure out how to combine bytes for different languages. Recently, there has been renewed interest in character-based models that adopt a byte tokenization scheme \citep{canine, byt5, charformer}---tokenization of texts into UTF-8 byte tokens. Although these byte-based models have shown competitive performance to subword models on multilingual NLU benchmarks~\citep{hu2020xtreme}, their performance on multilingual generation, especially on multilingual NMT is still underexplored. Although \citet{shaham-levy-2021-neural} recently demonstrate the effectiveness of byte tokenization for bilingual machine translation, a comprehensive study of such byte-based methods on the multilingual paradigm across a wide variety of languages and domains is still missing. Particularly in the multilingual setting, as characters in different languages can be tokenized into a varying number of byte tokens, this produces byte sequences much longer than the original sentences, and vanilla byte-based MT models can only \textit{implicitly} reconstruct character-/word-level representations from byte tokens in an entirely data-driven fashion.


To remedy these issues, we propose two variants of \textbf{Lo}cal \textbf{B}yt\textbf{e} \textbf{F}usion (LOBEF) \footnote{The code can be found at \url{https://github.com/makeshn/LOBEF_Byte_NMT}} techniques that explicitly aggregate byte tokens to learn character-/word-level representations for byte-based NMT models. Our first variant utilizes four $n$-gram convolutional layers to aggregate bytes for learning character-level information, and our second variant utilizes word boundaries to aggregate a span of bytes for learning word-level context. 
We conduct extensive experiments to compare our methods with the vanilla byte-based model and the embeddingless model from \citet{shaham-levy-2021-neural} in a multilingual translation setting. Our many-to-one translation results show that aggregating local information in earlier layers encourages the model to capture local information for seven source languages, yielding an average gain of up to 1.4 BLEU over the vanilla byte-based NMT model while performing competitively with subword models. We further demonstrate the effectiveness of LOBEF on the zero-shot/few-shot cross-lingual transfer and cross-domain adaptation settings, showing the flexibility of byte-based NMT models over subword baselines when fine-tuning is required for data adaptation. Additionally, our method also improves over vanilla byte-based NMT models for adaptation. Our contributions are as follows:
\begin{itemize}[leftmargin=10pt]\itemsep-0.2em 
    \item To the best of our knowledge, we are the first to evaluate byte-based embeddingless NMT models in a multilingual translation setting.
    \item To further improve the encoding of local semantics for byte-based NMT, we propose two variants of local fusion techniques based on character-/word-level aggregation over byte tokens.
    \item We provide a fine-grained analysis to show the effectiveness of byte-based models on cross-lingual and domain adaptation settings.
\end{itemize}

\section{Preliminaries}
\label{sec:prelim}


\subsection{Unicode and UTF-8}
Unicode
is a universal, platform-agnostic standard for handling text in most of the world's writing systems, covering characters in all of the world's living languages as well as emoji and non-visual codes. Each code point defined by Unicode is mapped to a unique integer, ranging from 0 to 10FFFF$_{16}$. For instance, the English character set \textit{A-Z} is denoted by the integers from \textit{97-122}. In modern computers, each Unicode code point can be implemented as bytes by multiple encoding protocols, and UTF-8 is the dominant encoding protocol used by over 95\% of webpages. 

In UTF-8, each Unicode code point is represented as one to four bytes (8 bits per byte) depending on the range of its Unicode integer. Some languages may have a combination of characters that require a varying number of bytes. For example, most characters in German require only a single byte, while some special characters like $\grave{a}$ or $\hat{a}$ use two bytes. Since the Unicode and the UTF-8 encoding scheme is already well-defined, we do not have to construct source and target vocabularies similar to how it is done for subword models. Tokenization and de-tokenization for byte based models is as simple as a single line of code in Python and does not involve any heuristic preprocessing.  In this paper, we adopt the UTF-8 byte tokens as inputs to our model. 
\subsection{Byte-based NMT}
\label{sec:byte_nmt}
\citet{shaham-levy-2021-neural} recently propose an \textit{embeddingless} NMT model that takes sequences of UTF-8 byte tokens as the inputs and outputs, and uses a fixed one-hot representation for each byte token instead of a dense learnable embedding vector. Such a byte-based NMT model eliminates the input and output token embedding layers usually used in subword-based NMT models, leading to a significant reduction in model parameters. 

Formally, given a source-target sequence pair from a parallel corpus $(x, y)\sim\Dcal$ where $x=(x_1, ..., x_N)$ and $y=(y_1, ..., y_M)$ are both sequences of byte tokens, the input sequence is first embedded by one-hot representations, i.e., $\text{Embed}(x)=\Xb\in\RR^{N \times d}$, and further encoded into the source hidden representation $\Zb$ by a vanilla $L$-layer Transformer encoder:
\begin{align}
    \Zb = \text{Encoder}(\Xb, L).
\end{align}
Finally, an attention-based decoder performs the attention over $\Zb$ and estimates the probability of predicting the next byte token $y_t$ by
\begin{align}\label{eq:decoder}
    P(y_t|y_{<t}, x) = \text{Decoder}(y_{<t}, \Zb).
\end{align}
Compared to subword-based NMT models, byte-based NMT models have shown effectiveness on bilingual machine translation, while their performance in multilingual machine translation is still unexplored. Especially in the many-to-one translation, the encoder is used to encode multiple languages which aggregate varying numbers of byte tokens (i.e., 1 to 4 bytes) to represent one character.

\section{Local Byte Fusion}
\label{sec:methods}
For languages that do not exclusively use the English character set, encoding them often requires more than one byte. Vanilla byte-based models can only \textit{implicitly} aggregate character-level or word-level representations for these languages, potentially resulting in poor interpretability and sub-optimal results in multilingual settings. Hence, we propose two fusion techniques that encourage models to \textit{explicitly} aggregate character-level and word-level information from byte sequences. 

We also adopt byte sequences as inputs and outputs for our model, and use vanilla Transformer as the backbone. 
As we focus on multilingual encoding in this work, we only modify the encoder, and adopt the same decoder architecture from \citep{shaham-levy-2021-neural}. Note that a more sophisticated design of the decoder will also involve a special design of decoding algorithms~\citep{libovicky-etal-2022-dont} which goes beyond the scope of this work. Besides, to conduct a more comprehensive study, we also consider the case where we retain embedding layers for the encoder and decoder of the byte-based model. This implies that instead of one-hot representations for the byte sequences, we can learn dense vector representations. Since the vocabulary size of all byte tokens is 256, this does not amount to adding a significant number of extra parameters. 

\subsection{$n$-gram Convolutional Fusion ($n$CF)}
\label{sec:byte-convolution}

Before we explicitly aggregate the character-level information, we first encode the input byte sequence by a shallow encoder with $L_s$ Transformer layers, which allows the model to have a shallow access to the sentence context before local fusion. 
\begin{align}
    \Sbb = \text{Encoder}(\Xb, L_s)
\end{align}

Since characters can be represented as a combination of 1 to 4 bytes depending on the languages, we apply four different 1-D convolutional layers to aggregate the $n$-gram byte tokens where $n\in\{1,2,3,4\}$. Specifically, we define $\conv_{n}$ as the 1-D convolution layer with a kernel of size $n$ and stride ${n}$. We do right padding at the end of the byte sequence. Therefore, when we use a stride $n$ greater than 1, the length of the input byte sequence is reduced by a factor corresponding to the stride $n$. We then define the output from the $\conv_{n}$ layer by:
\begin{align}
    \left[\fb^{n}_{1}, \cdots, \fb^{n}_{N\over n}\right] \in \RR^{{N\over n}\times d} &\leftarrow \conv_{n} \left( \Sbb \right).
\end{align}



To make all of the outputs the same length as the input, we repeat the output tokens in place by a factor corresponding to the stride length $n$. 
\begin{align}
    \Fb^n = \left[\repeatt(\fb^{n}_{1}, n), \cdots, \repeatt\left(\fb^{n}_{ {N\over n}}, n\right)\right],
\end{align}
where $\repeatt(\xb, n)$ creates $n$ copies of a vector $\xb$.
Applying this repetition process to the output from each of the convolution layers, we have four representations of equal sequence length as the source sequence,\footnote{Extra tokens at the end are truncated to ensure equal length.} i.e., $\Fb^1, \Fb^2, \Fb^3, \Fb^4 \in \RR^{N\times d}$. 
We pass these representations through a linear layer to get a single weighted representation: 
\begin{align}
    \Fb = \sum_{n=1}^4 \lambda^n \Fb^n,
\end{align}
where $\mathbf{\lambda}=[\lambda^1, \cdots, \lambda^4]$ are weights for the $n$-gram representations. We pass this weighted representation to the remaining ($L-L_s$) Transformer layers to obtain the final encoder hidden representation which is further sent to the decoder by Eq.~(\ref{eq:decoder}).
\begin{align}\label{eq:remain_encoder}
   \Zb =  \text{Encoder}\left( \Fb,L-L_s \right)
\end{align}

The $n$-gram fusion enables the model to learn what combination of the input byte sequence representations results in better character-level features. 

\subsection{Word-based Self-attention Fusion (WSF)}
\label{sec:block-attention}

In addition, we also propose a word-based self-attention fusion method that utilizes the word boundary information in the raw sentence to aggregate byte tokens within the same word. As characters in most languages are represented by more than one byte and words contain varying number of characters, using byte tokens as input to the model results in a much longer sequence. Therefore, this property may require the model to recognize a meaningful span of byte tokens in order to capture the semantic of a word token in the raw sentence. However, vanilla byte-based NMT models (\secref{sec:byte_nmt}) use the traditional \textit{full self-attention}, which implies that every byte token in the sequence attends to all byte tokens even though some far-away byte tokens may have little association to the query byte. Besides, as words are represented by a span of bytes in a small vocabulary of size 256, it is likely to produce a high attention weight between two identical byte tokens even when these byte tokens are used in two completely irrelevant words.

We tackle this issue by aggregating local information of a byte span for a word using a \textit{block-wise} self-attention. Formally, for a byte sequence $x=(x_1,\cdots, x_N)$, we define its (untokenized) word sequence as $w=(w_1, \cdots, w_T)$ and a mapping $\pi: [t]\rightarrow [a:b]$ that maps the word index $t$ to the beginning and the end indices of the corresponding byte span, i.e., $w_t=x_{\pi(t)}=x_{a:b}$. By leveraging the word boundary, we naturally break the long byte sequence into a list of sub-sequences, then we apply an $L_w$-layer Transformer encoder to encode byte tokens only in their sub-sequences:
\begin{align} \label{eq:WSF}
    \Fb_{\pi(t)} = \text{Encoder}(x_{\pi(t)}, L_w), \forall t \in [1:T],
\end{align}
where $\Fb_{\pi(t)}\in\RR^{|b-a|\times d}$ is hidden representation of byte tokens in the $t$-th word spanning over the sub-sequence $x_{a:b}$. This allows byte tokens to effectively aggregate local information for each word token, which is useful for the model to distinguish identical byte tokens used in two different words. Note that the word-based self-attention in Eq.~(\ref{eq:WSF}) can be efficiently implemented by pre-computing a \textit{block-wise} attention mask matrix (Figure~\ref{fig:main-figure} (c)), ensuring that self-attenion is only performed among a byte span of a word in a Transformer layer. Finally we obtain the \textit{word-aware} representation of the input byte sequence $\Fb$ by putting $\Fb_{\pi(t)}$ in the word order, i.e., $\Fb=[\Fb_{\pi(1)},\cdots, \Fb_{\pi(T)}]\in\RR^{N\times d}$, and feed $\Fb$ as input to the remaining $(L-L_w)$ Transformer layers similar to Eq.~(\ref{eq:remain_encoder}). 

\section{Experimental Settings}
\label{sec:expriments}



\subsection{Datasets}

\paragraph{Multilingual Many-to-One Translation:} We use the OPUS public data~\citep{tiedemann-2012-parallel} to construct a multilingual parallel corpus that has a fair mix of high-resource and low-resource languages. We train a multilingual translation model from seven source languages to English. Table \ref{tab:multilingual-data} shows the statistics of the training data. We use the Flores-101~\citep{Goyal2022TheFE} benchmark to evaluate the performance of our models. 

\paragraph{Zero-shot Cross-lingual Translation:} Following \citet{neubig-hu-2018-rapid}, we use the same Ted Talk dataset that include four language pairs where each pair has a high-resource language (HRL) and a low-resource languages (LRL) written in the same script. Table \ref{tab:cross-lingual-data} shows the statistics of the dataset.

\paragraph{Cross-domain Adaptation:} In this task, we train all models on the WMT19 German-English dataset on the news domain, and directly evaluate on the test data used in \citet{aharoni-goldberg-2020-unsupervised} from three diverse domains (Koran, IT, Medical).

\subsection{Models}

To fully evaluate the efficacy of byte-based techniques, we consider models under settings where we learn \textit{dense embeddings} for the input byte tokens (\dense) as well as the \textit{embeddingless} case where there are no learnt embeddings (\embedless). Our main baseline for comparison is the vanilla byte-based model and the \embedless model proposed in \citet{shaham-levy-2021-neural}. We also include results of subword and character-based models for a holistic comparison.

\paragraph{Subword Model:}
We use BPE models trained using \texttt{Sentencepiece}\footnote{\url{https://github.com/google/sentencepiece}} as our subword model.  
\paragraph{Char Model:}
We use character-based models with inputs and outputs being character sequences. 

\paragraph{Byte Based Models:}
For each of these models, we consider both \embedless variants where models do not have learnt embeddings and \dense variants where we learn continuous dense embeddings.

\begin{itemize}[leftmargin=10pt]\itemsep-0.2em 
    \item \textbf{Byte}: Similar to the vanilla byte-based model proposed by \citep{shaham-levy-2021-neural}, the inputs and outputs are UTF-8 byte tokens. 
    \item \textbf{\byteconv}:  We use a shallow Transformer encoder\footnote{Appendix~\ref{sec:num-layer-conv} shows that \byteconv works best on \{deu,khm\}-eng translations when fusing lower-layer representations.} ($L_s=1$) and four convolutional layers to fuse character-level information and learn a weighted \textit{$n$-gram} representation (\secref{sec:byte-convolution})
    \item \textbf{\byteword}: We use a $L_w$-layer Transformer encoder\footnote{Appendix~\ref{sec:num-layer-word} shows that $L_w=4$ empiricall works best.} with a word-based self-attention over byte tokens within word boundaries (\secref{sec:block-attention}).
\end{itemize}

\begin{table}[t]
\small
\centering
\begin{tabular}{l@{\hskip 8pt}l@{\hskip 8pt}l@{\hskip 8pt}l@{\hskip 8pt}l@{\hskip 8pt}l}
\toprule
\textbf{Lang.} & \textbf{ID} & \textbf{Script} & \textbf{Fertility} &  \textbf{\#Train} & \textbf{\#Test}  \\
\midrule
German & deu & Latin & 1.6 &2.56M & 1,012    \\
Hindi & hin & Devanagari & 1.6 & \phantom{0}1.6M & 1,012 \\
Nepali & npi & 	Devanagari & 2.0&  \phantom{.}445K & 1,012  \\
Tamil & tam & Brahmic & 2.6 &\phantom{.}268K & 1,012  \\
Telugu & tel & Brahmic & 2.5 & \phantom{.}108K & 1,012 \\
Khmer & khm & Khmer &8.5 & \phantom{.}127K & 1,012    \\
Lao & lao & Lao & 9.5 & \phantom{0}2.7K     & 1,012  \\
\bottomrule
\end{tabular}
\vspace{-2mm}
\caption{Writing scripts, fertility of seven source languages, no. of sentences in the many-to-English training set from OPUS and test set from Flores-101.}
\label{tab:multilingual-data}
\vspace{-5mm}
\end{table}

\subsection{Multilingual Translation}
\label{sec:multilingual}

In this experiment, we evaluate the subword and byte-based models on many-to-one translation (\textbf{xx-eng}) where \textbf{xx} refers to seven source languages listed in Table \ref{tab:multilingual-data}. We first clean the training data by removing sentences that are longer than 800 bytes in either the source or the target side, and then tokenize the sentences using the \texttt{Moses} tokenizer.\footnote{\url{https://github.com/moses-smt/mosesdecoder}} Doing such preprocessing does not affect the diversity of the dataset in terms of length as less than 0.5\% of the samples are discarded. The byte-based models do not have any preprocessing apart from the \texttt{Moses} tokenization and even whitespaces are included as valid tokens. For low-resource languages that share the same script as high-resource languages, we can reuse the same tokenizer for the high-resource language. For the subword-based model, we construct a shared vocabulary of 64K BPE tokens for all the source languages and an English vocabulary of 8K BPE tokens for this experiment. All models are trained for the same number of epochs on our OPUS train set, and evaluated on the Flores-101 test set.

\subsection{Cross-lingual Transfer}
\label{sec:cross-lingual-transfer}
This experiment evaluates how effective subword and byte-based methods are in transferring performance across languages that share similar language scripts. We train both the subword and byte-based models on parallel data in a high-resource language (HRL) for 50K steps, and evaluate them in a zero-shot manner on the corresponding low-resource language (LRL) without training on any LRL data. Table \ref{tab:cross-lingual-data} shows the data statistics. We focus on \textbf{xx-eng} translation where \textbf{xx} is either HRL or LRL. 

In the case of subword models, this amounts to constructing a vocabulary (i.e., BPE tokenizer) based on only the HRL data and using that to tokenize the LRL data, while byte-based models use an universal tokenization scheme to tokenize both HRL and LRL data into UTF-8 byte tokens.

We also investigate a few-shot setting where the models pre-trained on the HRL data is further finetuned on a few parallel training samples in LRL. We examine the impact of different numbers (i.e., 1K, 2K, 3K, and 4K) of few-shot samples on the translation performance of these models. We finetune all models for 5K steps on the few-shot samples, and then evaluate them on the test set in LRL.

\begin{table}[ht]
\small
\centering
\resizebox{\columnwidth}{!}{\begin{tabular}{@{}l@{\hspace{1ex}}l@{\hspace{1ex}}c|l@{\hspace{1ex}}l@{\hspace{1ex}}c@|c@{\hspace{1ex}}c@{}}
\toprule
\multicolumn{3}{@{}c|}{\textbf{HRL}} & \multicolumn{4}{@{}c}{\textbf{LRL}} \\
\textbf{Language} & \textbf{ID} &  \textbf{Train Size} &  \textbf{Language} & \textbf{ID}  & \textbf{Test Size} & \textbf{<unk>\%}\\
\midrule
Turkish & tur & 182k & Azerbaijani & aze & \phantom{0.}1k & 41.5\% \\
Russian & rus & 208k & Belarusian & bel & 0.6k & 48.1\%\\
Portugese & por & 185k & Galician & glg & \phantom{0.}1k & 23.7\%\\
Czech & ces & 182k & Slovak & slk & 2.5k & 30.0\%\\
\bottomrule
\end{tabular}}
\vspace{-2mm}
\caption{Sentence sizes for LRL/HRL, and unknown token rate on the LRL test set using HRL BPE tokenizers}
\label{tab:cross-lingual-data} 
\vspace{-6mm}
\end{table}

\begin{table*}[!htbp]
\begin{center}
\begin{small}
\resizebox{0.8\textwidth}{!}{%
\begin{tabular}{@{}ll|ccccc|ccc@{}}
\toprule
\multicolumn{2}{@{}c|}{\textbf{Lang. Pairs}} & \multicolumn{5}{c|}{\textbf{\dense Models}} & \multicolumn{3}{c}{\textbf{\embedless Models}} \\
\textbf{Src} & \textbf{Tgt} &  \textbf{Subword}  & \textbf{Char}  & \textbf{Byte}  & \textbf{\byteconv} & \textbf{\byteword} & \textbf{Byte}  & \textbf{\byteconv} & \textbf{\byteword} \\
\midrule

deu  & eng &   31.5 & 29.2 & 31.3 & \textbf{32.1} & 31.7 & 31.1 (-0.4) & \textbf{31.6} (-0.6) & 31.3 (-0.3)\\
hin  & eng &   24.8 & 21.9 & 23.9 & \textbf{25.5} & 25.4 & 24.3 (-0.2) & \textbf{25.6} (+0.1) & 24.9 (+0.0)\\
npi  & eng &  18.1 & 17.3 & 18.2 & 19.8 & \textbf{19.9} & 17.9 (-0.1) & 19.1 (-0.7) & \textbf{19.2} (-0.8)\\
tam  & eng &  18.2 & 17.4 & 17.9 & \textbf{19.5} & 18.9 & 18.3 (-0.4) & \textbf{19.1} (-0.4)  & 18.8 (-0.1)\\
tel  & eng &   20.3 & 17.5 & 20.5 & \textbf{22.2} & 22.1 & 19.9 (-0.5) & 21.1 (-1.1) & \textbf{20.8} (-1.0)\\
khm  & eng &   12.6 & 11.1 &  12.4 & 13.5 & \textbf{13.9} & 12.2 (-0.2) & \textbf{12.6} (-1.3) & 13.1 (-0.8) \\
lao  & eng &   \textbf{9.2} & \phantom{0}7.7 & \phantom{0}5.9 & \phantom{0}6.4 & \phantom{0}6.5 & \phantom{0}5.8 (+0.1) & \phantom{0}6.3 (-0.1) & \phantom{0}\textbf{6.9} (+0.4)\\ \midrule
\multicolumn{2}{@{}c|}{\textbf{Avg.}} & 19.2 & 17.4 & 18.6 & 19.9 & 19.7 & 18.5 (-0.1) & 19.4 (-0.5) & 19.3 (-0.4) \\ 
\bottomrule
\end{tabular}}
\end{small}
\end{center}
\vspace{-4mm}
\caption{BLEU scores of the \dense and \embedless models on the Flores-101 dataset. Highest scores for each language pair on these two sets of models are highlighted in bold font. The differences of BLEU scores between \embedless models and their corresponding \dense variants are highlighted in the brackets. \label{tab:multilingual-results}} 
\vspace{-4mm}
\end{table*}

\subsection{Cross-domain Adaptation}

Having translation systems adapt to domains apart from the one it has been trained on is a good measure of how robust models are. In the experiment, we compare subword and byte-based models on how effectively they translate sentences from domains that are not part of the training set. Similar to the cross-lingual transfer setting (\secref{sec:cross-lingual-transfer}), we train both subword and byte-based models on the source domain (News) and evaluate them in a zero-shot manner on three target domains (Koran, IT, Medical). Each model is trained on the source domain dataset for 50K steps, and then evaluated on each of the target domain test sets. 

\subsection{Hyperparameters}

We use the \texttt{Fairseq}\footnote{\url{https://github.com/facebookresearch/fairseq}} library as the codebase. To make a fair comparison, we strictly follow the architectural choice of \citet{shaham-levy-2021-neural} and employ the vanilla transformer encoder-decoder architecture as our backbone for all experiments. For all models, we use a total of 6 Transformer layers for the encoder and 6 layers for the decoder with 8 attention heads, 512 hidden units and the feed-forward dimension of 2048. We use the Adam\citep{adam} optimizer with an inverseq square root learning rate scheduler, and warm up 4K steps to reach a peak learning rate of 5e-4. We apply a weight decay of 1e-4 and a label smoothing of 0.1. We also train all models for an equal number of epochs in all the experiments. 

\subsection{Evaluation}
For a fair, consistent evaluation, we follow \citet{shaham-levy-2021-neural} in using \texttt{Sacre-BLEU}\footnote{\url{https://github.com/mjpost/sacrebleu}} with 13a tokenizer for all language pairs using the raw text to compute the BLEU scores.

\section{Results}
\label{sec:results}
 
In this section, we detail the results of the various experiments and discuss their implications.

\subsection{Multilingual Translation}
\label{sec:translation-results}
Table~\ref{tab:multilingual-results} shows the BLEU scores on the test set of the FLORES-101 data for many-to-one translation. We further investigate the following questions. 

\paragraph{Do we need dense embeddings?} In line with the findings of  \citep{shaham-levy-2021-neural} that embeddingless models are competitive with subword models in bilingual settings, we find that they perform on par with their corresponding models that use dense embeddings with an average difference of less than 0.5 BLEU over seven languages. We find that for six out of the seven source languages under consideration, the byte-based models perform competitively with the subword models. However, subword models still hold the edge for extremely low-resource languages such as \textit{Lao-English} translation with only 2.7K training data. Besides, as Lao's written script is not shared with any of the other languages, we hypothesize that training byte-based multilingual models requires more training data in order to figure out the different fusion of byte tokens across languages while subword models with an explicit vocabulary for all languages do not have this requirement. 

\paragraph{How effective is character/word fusion?} Our proposed methods (\byteconv \ and \byteword) that induce higher level semantic representations for bytes improve over vanilla byte-based models in both cases (\embedless and \dense models) on all language pairs with an average gain of up to 1.4 BLEU. Since sequence lengths tend to be extremely long when using byte sequences, aggregating information locally in the lower layers enables the model to quantitatively obtain higher scores than even subword-based models except in the extremely low-resource regime. 

\paragraph{Which fusion works better?} Comparing our two proposed variants, the \byteconv model performs slightly better than the \byteword model in the \dense case, while both perform comparably in the \embedless case. 
In particular, \byteconv performs better than \byteword on relatively high-resource languages (e.g., German, Hindi) with more than 1M training data. Besides, both variants perform comparably on low-resource languages (e.g., Khmer, Lao) with large fertility scores.

\subsection{Cross-lingual Transfer}
\label{sec:transfer-results}

The performance of byte-based and subword models on cross-lingual transfer is shown in Table \ref{tab:cross-lingual-results}. As \byteword and \byteconv have shown comparable performances in Table~\ref{tab:multilingual-results}, we only include the \byteconv variant in the comparison below.

\paragraph{Does universal tokenization work?} When evaluating subword and byte-based models in a zero-shot setting (\secref{sec:cross-lingual-transfer}), byte-based models outperform subword baselines by a clear margin of up to 6.1 average BLEU over all languages. The gains compared to vanilla byte baseline is 1.2 BLEU for \dense variant and 0.7 BLEU for \embedless models. Our results indicate that even for languages written in the same script, a rigid subword schedule is infeasible for NMT models to perform an effective cross-lingual transfer. Particularly, we observe a significant increase in BLEU in the glg-eng and slk-eng translations when using byte tokens as inputs.

\paragraph{Does fusion help cross-lingual transfer?} We find that using the \byteconv fusion variant leads to marginal improvement over the vanilla byte-based model with an average gain of up to 0.4 BLEU. It should be noted that most of these language pairs share the same script and hence the convolution fusion technique works very well. Investigating whether such fusion techniques work for languages that do not share the same script can be explored in future work. 

\begin{table}[t]
\begin{center}
\begin{small}
\resizebox{\columnwidth}{!}{%

\begin{tabular}{@{}cc|c@{\hspace{1ex}}c@{\hspace{1ex}}c@{\hspace{1ex}}|c@{\hspace{1ex}}c@{}}
\toprule
\multicolumn{2}{@{}c|}{\textbf{Lang. Pair}} & \multicolumn{3}{c}{\textbf{\dense Models}} & \multicolumn{2}{c}{\textbf{\embedless Models}} \\
\textbf{Src} & \textbf{Tgt} &  \textbf{Subword}  & \textbf{Byte}  & \textbf{\byteconv} & \textbf{Byte}  & \textbf{\byteconv}  \\
\midrule

aze & eng & 3.7 & 6.9 & \textbf{7.8} & 5.7 & 6.9 \\
bel & eng & 1.7 & 3.9 & \textbf{5.3} & 4.6 & 5.4 \\
glg & eng & 7.6 & 15.2 & 16.7 & 16.4 & \textbf{17.2} \\
slk & eng & 2.9 & 11.4 & \textbf{12.6} & 11.9 & 12.2 \\ 
\midrule 
\multicolumn{2}{@{}c|}{\textbf{Avg.}} & 4.0 & 9.4 & \textbf{10.6} & 9.7 & 10.4\\
\bottomrule
\end{tabular}}
\end{small}
\end{center}
\vspace{-3mm}
\caption{BLEU scores of the \dense and \embedless models on the Ted Talk dataset. Highest score among all models is in bold font.} 
\label{tab:cross-lingual-results}
\vspace{-4mm}
\end{table}

\begin{figure}[th]
    \centering
    \includegraphics[width=\linewidth]{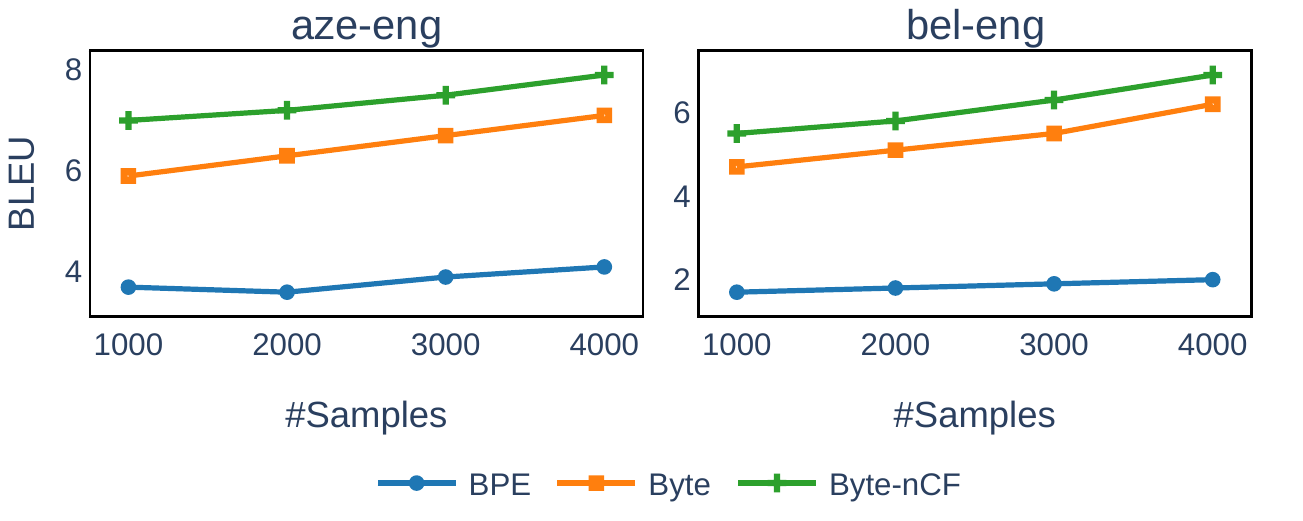}
    \vspace{-3mm}
    \caption{Few shot translation performance of BPE and Byte (\embedless) based models. }
    \label{fig:few-shot-aze-bel}
    \vspace{-4mm}
\end{figure}

\paragraph{Does the few-shot setting improve performance?} Figure \ref{fig:few-shot-aze-bel} shows the translation performance in terms of BLEU for the BPE and byte-based models in the few-shot setting. We find that as the number of training data in LRL increases, the performance of the byte-based models improves, and \byteconv consistently improves over the vanilla byte model. The BPE baseline suffers from the issue of having a high unknown token rate and cannot take full advantage of the additional training data.  

\subsection{Cross-Domain Adaptation}
\label{sec:domain-adaptation-results}

Table \ref{tab:cross-domain-results} shows the results of subword and byte-based models on zero-shot cross-domain adaptation. The first row indicates the BLEU scores on the \textit{in-domain} test set (WMT19 News), and the other rows showcase the performance on \textit{out-of-domain} test sets (Koran, IT, Medical). 

\paragraph{Are byte-based models robust to domain shift?} We find that the performance of both subword and byte-based models is susceptible to domain shifts. The BLEU scores on other domains are significantly lower for all variants. However, on comparison, we find that byte-based models are more robust than subword models against domain, yielding higher BLEU scores on all out-of-domain test sets. 

\begin{table}[]
\begin{center}
\begin{small}
\resizebox{\columnwidth}{!}{%
\begin{tabular}{@{}c|ccc|cc@{}}
\toprule
\multicolumn{1}{@{}c}{\textbf{Domain}} & \multicolumn{3}{c}{\textbf{\dense Models}} & \multicolumn{2}{c}{\textbf{\embedless Models}} \\
& \textbf{Subword}  & \textbf{Byte}  & \textbf{\byteconv} & \textbf{Byte}  & \textbf{\byteconv}  \\
\midrule

WMT19 News & 17.6 & 21.2 &21.3  & 21.1 & \textbf{21.5}   \\
Koran & 1.8 & 6.6 & \textbf{7.4} & 6.8 & 7.7  \\
IT & 3.9 & 10.4 & \textbf{11.6} & 10.2 & 11.3  \\
Medical & 4.1 & 13.6 & 15.3   & 12.9  &  \textbf{15.4}  \\
\bottomrule
\end{tabular}}
\end{small}
\end{center}
\vspace{-3mm}
\caption{BLEU scores of the \dense and \embedless models on zero-shot cross-domain adaptation.} 
\label{tab:cross-domain-results}
\vspace{-3mm}
\end{table}

Employing convolution fusion with the byte-based models improves performance over subword-based models, especially in the IT and medical domains. The issue with cross-domain adaptation remains that each new domain consists of specific jargon and entities that are not captured in the source domain. This inhibits the models from capturing the required semantic information to translate out-of-domain sentences effectively.

\begin{figure*}
    \centering
    \begin{subfigure}{0.35\textwidth}
        \includegraphics[width=\linewidth, height=4cm]{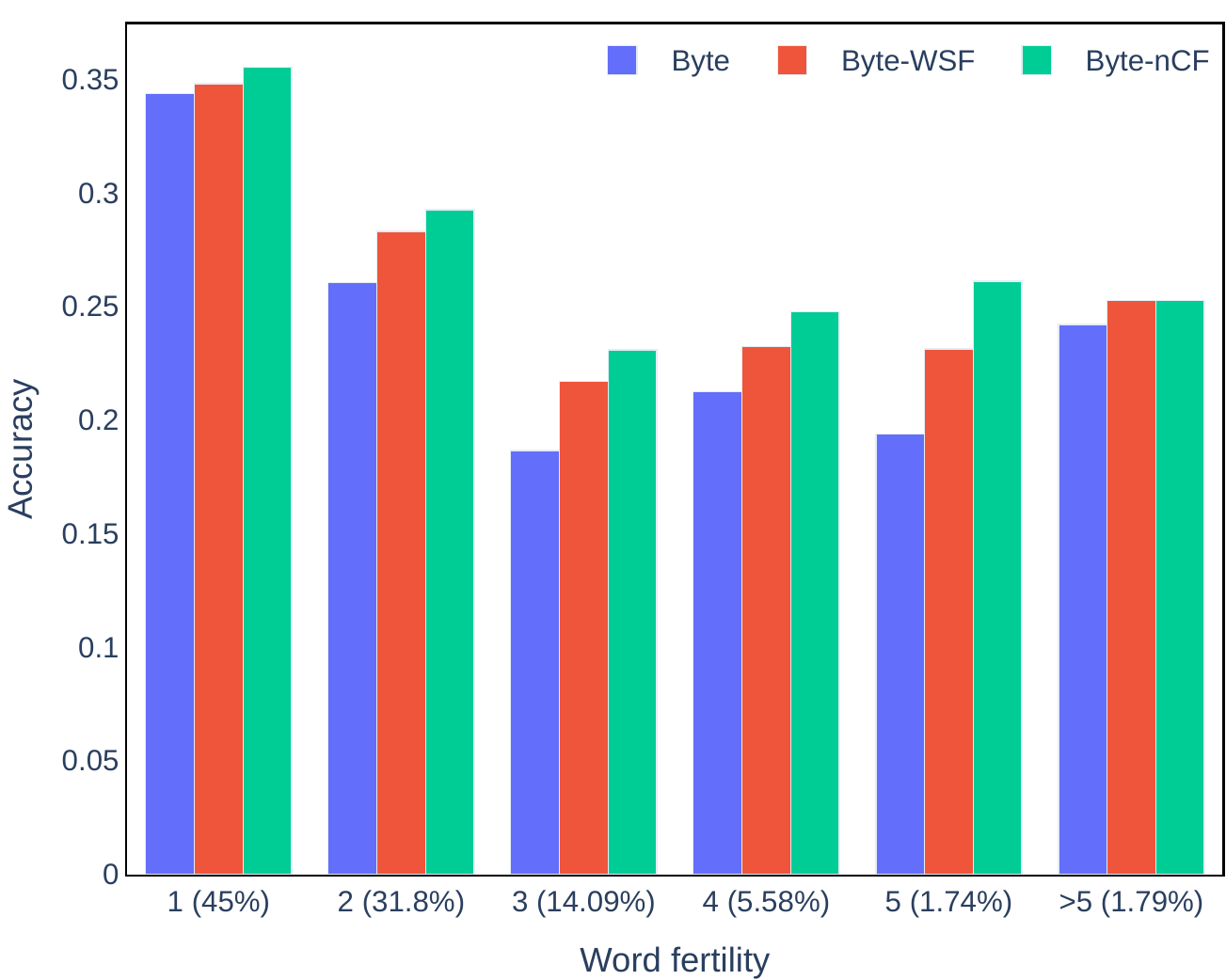}
          \caption{Fertility Score vs Word accuracy}
          \label{fig:fertility-recall}
      \end{subfigure}
    \begin{subfigure}{0.35\textwidth}
        \includegraphics[width=\linewidth, height=4cm]{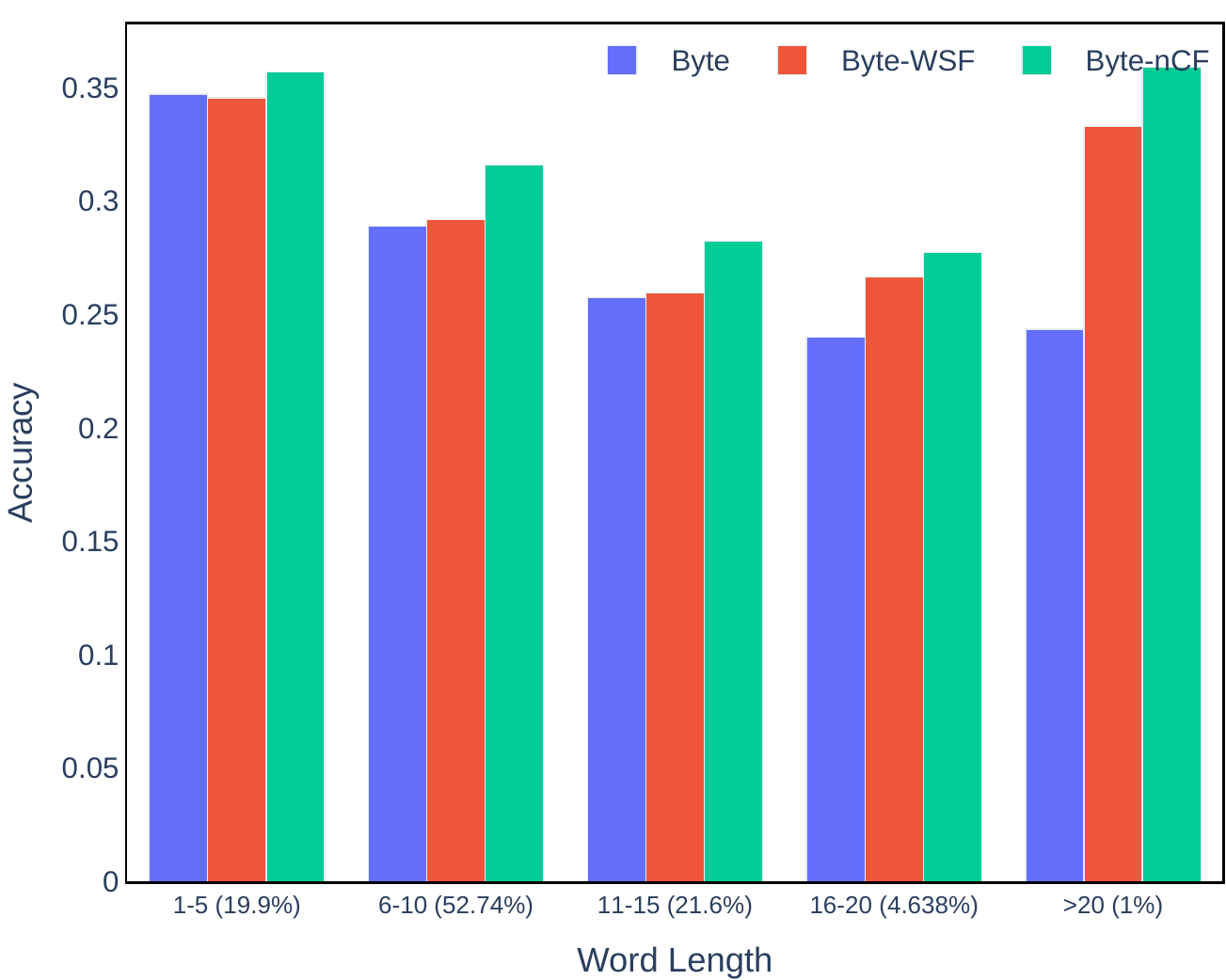}
          \caption{Word Length vs Word accuracy}
          \label{fig:length-recall}
      \end{subfigure}
    \vspace{-3mm}
    \caption{Translation Word Accuracy grouped by word fertility/length for \embedless Byte models on deu-eng. }
    \label{fig:selection}
    \vspace{-6mm}
\end{figure*}

\section{Discussion and Analysis}
\label{sec:analysis}
Next, we further present a qualitative analysis of byte-based models and our proposed variants. 

We use the \texttt{compare-mt} toolkit \citep{compare-mt} to holistically analyze how the outputs of these models differ and what aspects different models excel at. We compare the multilingual NMT models (\secref{sec:translation-results}) on the \textit{German-English} translation as a sample language pair for the analysis. Specifically, we group all source words in the test sentences into buckets by the word fertility score (Figure \ref{fig:fertility-recall}) and the word length in terms of characters (Figure \ref{fig:length-recall}). Recall that the word fertility score measures how many subword units a word is broken into, and we use the BPE tokenizer used in Section \ref{sec:translation-results}. We evaluate byte-based models (i.e., Byte, \byteword, \byteconv) using one-hot representations on each group in terms of source word translation accuracy. 

 As German is a high-resource language, most German words (45\%) have a fertility score of 1 implying that they are not segmented by the BPE tokenizer, and all byte-based methods perform comparable on these words. We find that the \byteconv method performs better than the other two byte-based methods on oversegmented words (as indicated by the accuracy on words with fertility scores above 4). We also find that the \byteconv method outperforms other methods on translating long words (depicted by the accuracy on words with length greater than 15 characters). Comparing to the word-based (\byteword) or sentence-level full self-attention (Byte), we hypothesize that this is a result of encoding a smaller sequence length when using the convolution fusion operation, reducing the pressure of byte-based models to capture too much information from a span of byte tokens. 

\section{Related Work}
\label{sec:related}

\paragraph{Subword Models} Byte Pair Encoding \citep{sennrich-etal-2016-neural}, Wordpiece \citep{wu-wordpiece} and SentencepPiece \citep{kudo-richardson-2018-sentencepiece} are widely-used subword tokenization schemes for NMT models, or perhaps most neural NLP models. However, the rigid tokenization scheme poses challenges in terms of oversegmenting low-resource languages and adapting a pre-trained model to new languages or new domains of different corpora~\citep{sun,bostrom-durrett-2020-byte, provilkov-etal-2020-bpe, kudo-2018-subword, godey-etal-2022-manta}. 


\paragraph{Character-level Models} Applying neural models directly on character sequences has been extensively studied \citep{sutskever,graves-lstm, kalchbrenner, zilly, melis2018on, Al-Rfou_Choe_Constant_Guo_Jones_2019, characterawareneural, gao-etal-2020-character, tay2022charformer}. \textit{Character-aware} methods were mainly developed by the use of word boundaries and convolutions over characters \citep{kim, jozefowicz, peters, el-boukkouri-etal-2020-characterbert,ma-etal-2020-charbert}. However, for machine translation, character-based NMT models~\citep{lee-etal-2017-fully,cherry-etal-2018-revisiting,libovicky-etal-2022-dont} still suffer from a high computational overhead of encoding and decoding much longer sequences. 

\paragraph{Tokenization-free Methods} Recent attempts have focused on using Unicode or UTF-8 encoding scheme to remove pre-built subword tokenizers from preprocessing. These byte-based methods have achieved promising results in terms of accuracy and speed-up in several multilingual language understanding tasks~\citep{canine, charformer, byt5} or bilingual translation tasks~\citep{shaham-levy-2021-neural}, while their application to multilingual or cross-lingual/domain settings is still underexplored. 

\section{Conclusion}
\label{sec:conclusion}

We propose a \textbf{Lo}cal \textbf{B}yt\textbf{e} \textbf{F}usion (LOBEF) method with two variants---one employing convolutions on byte $n$-grams and the other utilizing word-based self-attention restricted to word boundaries. We show that these two fusion variants improve upon vanilla byte-based models indicating that neural machine translation models benefit from the explicit aggregation of local semantic information for characters or words at lower layers of neural networks. Our experiments show that both \embedless and \dense versions of byte-based models perform competitively on multilingual machine translation and even beat subword baselines on multiple language pairs. We also conduct an investigation of the effectiveness of byte-based techniques in both zero-/few-shot cross-lingual transfer and domain adaptation settings, and find that they outperform subword models by a large margin. 


\section{Limitations}
\label{sec:limitations}
Despite achieving high translation performance on various language pairs, LOBEF has some limitations, coming from the nature of processing UTF-8 byte sequences.

\paragraph{Speed:} As shown in Table~\ref{tab:time-analysis} in the appendix, the inference times for byte-based models are higher when compared to subword-based models. It is also worth noting that we use the same amounts of model parameters for a total of 6 Transformer encoder layers and 6 Transformer decoder layers for all models in comparison. As shown in Table~\ref{tab:parameter-analysis}, byte-based models can effectively reduce the amounts of parameters for the embedding layers comparing to the subword-based models, leading to faster training time as shown in Table~\ref{tab:time-analysis}. However, as indicated by \citet{byt5}, by adding more encoder layers, we can construct byte-based models with comparable amounts of parameters as subword-based models, and these larger byte-based models still require much longer time for training than subword-based models.

\paragraph{Extremely Low-resource Languages:} The performance of byte-based models on extremely low-resource languages (e.g., 2.7K training data for Lao-English) is still lower than subword models especially in the multilingual setting. We suspect that byte-based methods require a relatively larger number of training data in order to aggregate information from a combination of byte tokens, comparing to subword-based models that explicitly maintain a subword vocabulary.



\paragraph{Extra Preprocessing:} The \byteword model requires an extra preprocessing step that pre-computes the attention mask corresponding to the words in each sentence. This adds a slight overhead before training, while training the \byteword model is as fast as the Byte model, as both model use the same Transformer architecture. However, for languages (e.g., Chinese) that do not have white-spaces to indicate the word boundary, we may rely on an off-the-shell word segmentation tool to preprocess the text.


\bibliography{anthology,custom}
\bibliographystyle{acl_natbib}

\clearpage
\newpage
\onecolumn
\appendix
\section*{Appendix}
\label{sec:appendix}

\section{Datasets used for Multilingual Training}

\begin{table}[htbp]
\begin{center}
\begin{tabular}{@{}ll@{}}
\toprule
Language & OPUS Corpora                                                         \\ \midrule
deu-eng  & Wikipedia, WMT-News, Bible                                           \\
hin-eng  & IITB Hindi-English corpus                                            \\
khm-eng  & CCAligned, GlobalVoices, QED, GNOME, TED2020, KDE4, tico-19, Tatoeba \\
lao-eng  & Wikimedia, TED2020, QED, GNOME, Ubuntu, Tatoeba                      \\
tel-eng  & Wikimedia, TED2020, QED, GNOME, Bible                                \\
tam-eng  & Wikimedia, TED2020, QED, GNOME, Tanzil                               \\
nep-eng  & Wikimedia, TED2020, QED, GNOME, Bible, GlobalVoices                  \\ \bottomrule
\end{tabular}
\end{center}
\caption{List of datasets from OPUS we use to construct our corpus for multilingual experiments.}
\end{table}

\section{Computational Efficiency} 
For comparing the byte based and subword models in terms of the number of parameters, training and inference times we consider the transformer base architecture. For the subword baseline, we consider a source vocabulary of 32k and 8k for the target vocabulary (English).

\paragraph{Parameters} When comparing the number of parameters in subword and byte based models, we find that byte based models have far fewer parameters (\textasciitilde30\% fewer) as compared to the subword baselines. We highlight the differences in Table \ref{tab:parameter-analysis}.

\begin{table}[h]
\begin{center}
\begin{small}
\resizebox{0.35\columnwidth}{!}{%
\begin{tabular}{@{}lll@{}}
\toprule
                                           & Model            & \#Parameters \\ \midrule
\multirow{4}{*}{\embedless}               & Byte             &    44.1M           \\
                                           & \byteconv &  46.5M              \\
                                           & \byteword &  44.1M              \\
                                                
\midrule
\multirow{5}{*}{\dense}                 & Subword & 68.7M \\  
                                            & Byte             &      44.3M        \\
                                           & \byteconv &    46.7M          \\ 
                                           & \byteword &    44.1M          \\ \bottomrule
\end{tabular} }
\end{small}
\end{center}
\caption{Comparison of number of parameters in subword and byte based models. We find that byte based models have on average 30\% fewer parameters than comparable subword models.}
\label{tab:parameter-analysis}
\end{table}

\paragraph{Training and Inference times} For comparing the training times, we train the models on the WMT19 de-en data for 5k steps. We use a warmup of 1k steps to eliminate any hardware discrepancies like GPU\footnote{All numbers are obtained using a single RTX 3090 GPU using a batch size of 7k tokens and 8 gradient accumulation steps for training.} cold starts. Since the byte based models are smaller than comparable subword models, they are ~20\% faster to train. The inference times are based on evaluating the model on the validation set across all batches\footnote{beam size of 3 and batch size of 256}. We find that byte based models are significantly slower than subword models for inference. The byte sequences are significantly longer than subword sequences and thus the decoding time takes a hit.  Table \ref{tab:time-analysis} shows the training and inference times for the subword and byte baesd models.

It should be noted that while the training time is shorter for byte-based approaches when comparing the same number of gradient steps, when we consider the training time for the same number of epochs, we do not observe faster training performance. Since byte sequences are much longer than subword sequences, training them for the same number of epochs involves using a longer number of training steps which makes their training times comparable.

\begin{table}[h]
\begin{center}
\resizebox{0.5\columnwidth}{!}{%
\begin{tabular}{@{}llll@{}}
\toprule
                                           & Model           & Train time(s) & Inference time(s) \\ \midrule
\multirow{2}{*}{\embedless}               & Byte              & 4184.1  & 50.963     \\
                                           & \byteconv   & 4387.6   & 52.315         \\
\midrule
\multirow{1}{*}{\dense} & Subword & 5176.8 & 22.445\\ \bottomrule 
\end{tabular}}
\caption{Comparison of the training time and inference time of subword and byte models. Byte-based models are faster to train, but are slower during inference than subword models.} 
\label{tab:time-analysis}
\end{center}
\end{table}

\paragraph{Computing Infrastructure}
All models are trained on a Linux server with 4 RTX 3090 GPUs and 16 CPU cores. On average, training all models on 2 GPUs for 200K steps can be finished within 24 hours. After training, we pick the best checkpoints based on the performance on the development set.

\section{Number of Shallow Encoding Layers for \byteconv}
\label{sec:num-layer-conv}

\begin{table}[H]
\begin{center}
\begin{tabular}{ccccccc}
\toprule
        \#Layer & 0 & 1 & 2 & 3 & 4 & 5 \\ \midrule
        deu-eng & 19.4 & \textbf{21.5} &21.4 &20.1 &20.4 & 19.7 \\
        khm-eng & 10.4 & \textbf{12.6} & 12.3 & 11.8 & 11.3 & 10.7 \\ \bottomrule
\end{tabular} 
\caption{BLEU score of \byteconv using $L_s$ shallow encoding layers. } 
\label{tab:num-layer-conv}
\end{center}
\end{table}

\section{Number of Word-based Self-Attention Layers for \byteword}
\label{sec:num-layer-word}
\begin{table}[H]
\begin{center}
\begin{tabular}{ccccccc}
\toprule
        \#Layer &  1 & 2 & 3 & 4 & 5 \\ \midrule
        deu-eng &   28.3 &27.9 &28.8 & 28.2& 27.4 \\ \bottomrule
\end{tabular} 
\caption{BLEU score of \byteword using $L_s$ word-based self-attention layers. } 
\label{tab:num-layer-word}
\end{center}
\end{table}

\section{Byte-BPE Baseline}

There are some works exploring the use of BPE vocabulary on byte tokens \citep{bytebpe} to get the best of both worlds - i.e. we would not have the out-of-vocabulary issue since every character can be represented as one of the 256 byte tokens and we also make use of the advantages of subword tokenization scheme to reduce the sequence length and decoding time. For a more reasonable comparison with similar-sized byte-based models, we strictly follow the settings of \citet{wang2019neural}, using BBPE models and setting vocab size to 2K or 4K. From our results below, we find that our proposed methods for byte fusion (using 256 vocab size) are slightly better than BBPE with 4K vocab size, with an avg. gain of up to 0.8 BLEU.   
  
\begin{table}[h]
\centering
\begin{tabular}{lcc}
\toprule
\textbf{Language Pairs} & \textbf{BBPE 2K} & \textbf{BBPE 4K} \\
\midrule
deu-eng & 30.8 & 31.1 \\
hin-eng & 24.7 & 25.1 \\
npi-eng & 18.4 & 18.7 \\
tam-eng & 19.3 & 19.6 \\
tel-eng & 20.4 & 21.1 \\
khm-eng & 11.2 & 11.6 \\
lao-eng & 6.1 & 6.4 \\
\midrule
\textbf{Avg.} & 18.7 & 19.1 \\
\bottomrule
\end{tabular}
\caption{BLEU scores for BBPE}
\end{table}

Note that BBPE may fall back to using single bytes when dealing with new byte combinations in a new language or domain. We also run a cross-lingual transfer experiment by training the BBPE(4k) model on tur-eng and evaluating it on aze-eng in a zero-shot manner. We find that it gets a BLEU of 7.4, which is better than vanilla byte models (6.9) but worse than our proposed Byte-nCF (7.8 BLEU). This suggests that even though there is a byte fall-back in such models, a certain fraction of BPE tokens used in high-resource languages may not be used in the low-resource language, and the model still has to implicitly fuse the new byte tokens for low-resource language, similar to vanilla byte baseline. 

This baseline is more comparable to BPE and is not the main baseline for our consideration since we are focused on improving over the vanilla byte-based methods. 


\section{Seen vs Unseen words - Domain Adaptation}

To compare how well the models generalize across domains, we compute the word accuracy score based on whether the source words were observed or unobserved during the training stage. Since all the models are trained on the WMT News domain and evaluated on the IT and Medical domains, analyzing the word accuracy on unseen words reveals where the performance gain stems from. We see that the Byte-nCF model has much higher word accuracies on the unseen words when compared with the Byte and BPE models 

\begin{figure*}
    \centering
    \begin{subfigure}{0.47\textwidth}
        \includegraphics[width=\linewidth]{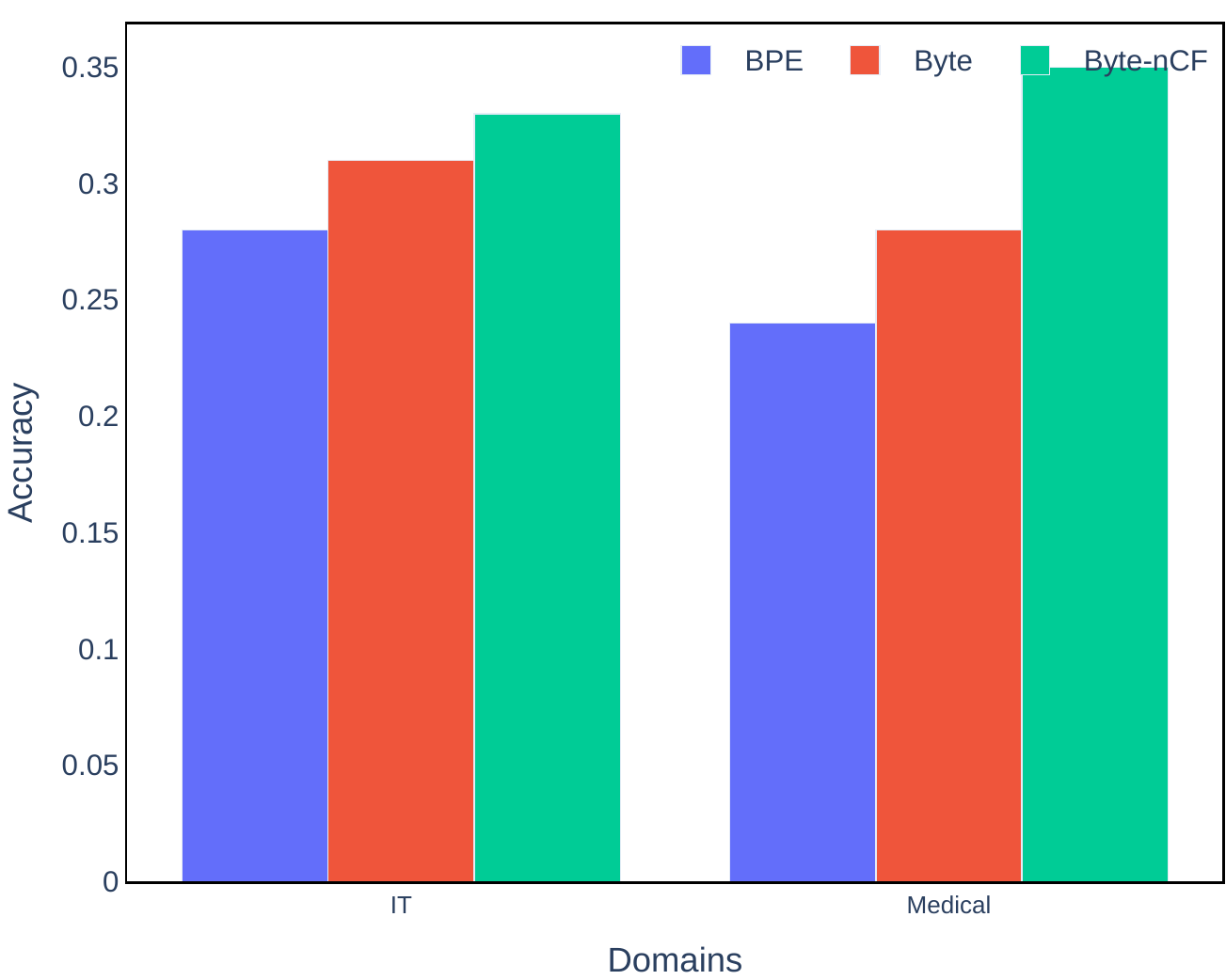}
          \caption{Word accuracy for seen words during training}
          \label{fig:fertility-recall}
      \end{subfigure}
    \begin{subfigure}{0.47\textwidth}
        \includegraphics[width=\linewidth]{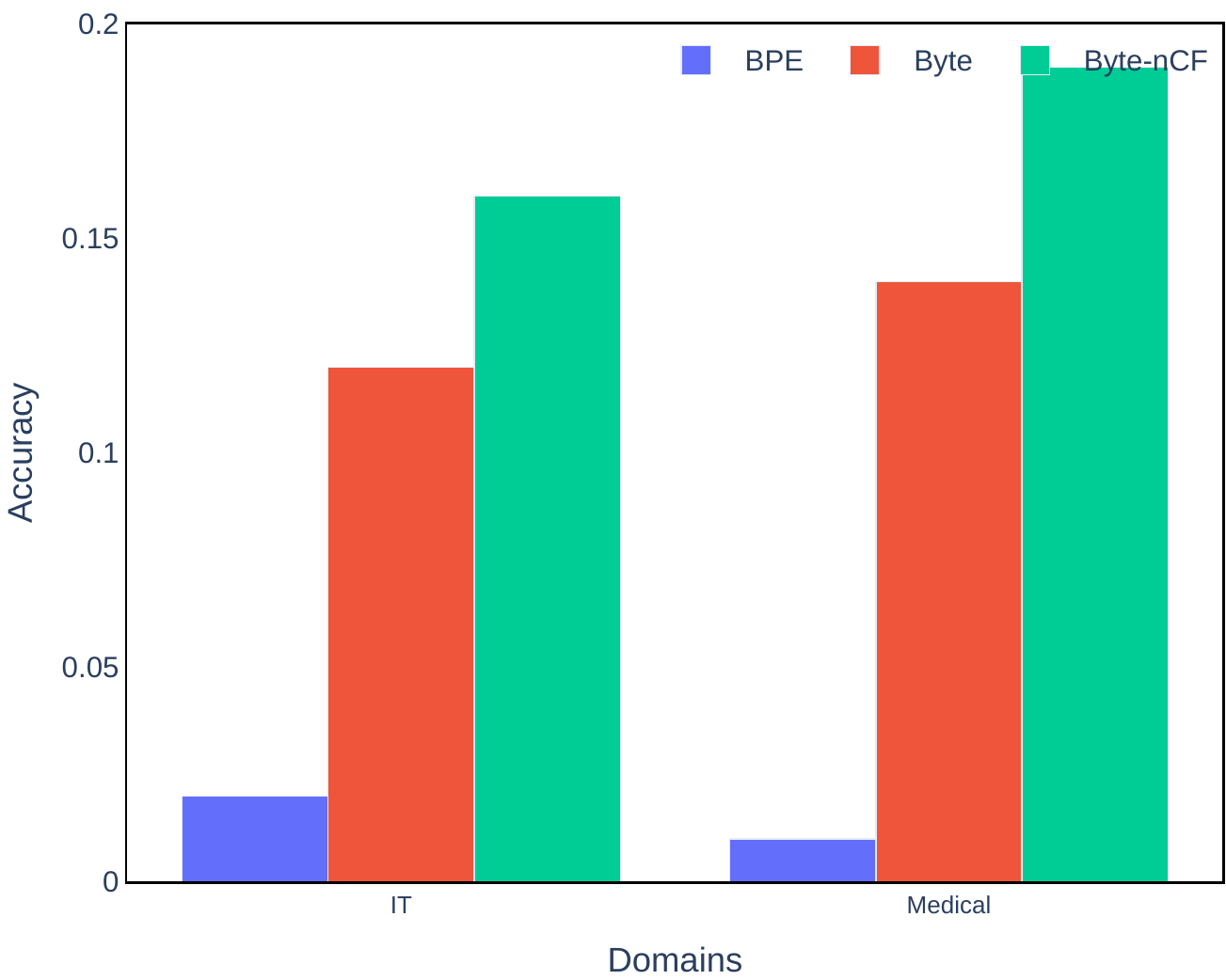}
          \caption{Word accuracy on unseen words during training}
          \label{fig:length-recall}
      \end{subfigure}
    \caption{Translation performance (indicated by Word Accuracy) grouped by whether the words were seen or unseen during training on different domains for \embedless Byte models on deu-eng. }s
    \label{fig:selection}
    \vspace{-6mm}
\end{figure*}


\end{document}